\pdfoutput=1
\documentclass[11pt]{article}

\usepackage[preprint]{acl}

\usepackage{times}
\usepackage{latexsym}

\usepackage[T1]{fontenc}
\usepackage[utf8]{inputenc}

\usepackage{microtype}

\usepackage{inconsolata}

\usepackage{listings}
\usepackage{tcolorbox}
\usepackage{times}
\usepackage{soul}
\usepackage{url}
\usepackage{graphicx}
\usepackage{amsmath}
\usepackage{amsthm}
\usepackage{booktabs}
\usepackage{algorithm}
\usepackage{algorithmic}
\usepackage{inconsolata}
\usepackage{import}
\usepackage{microtype}
\usepackage{layout}
\usepackage{tabularx, makecell}
\usepackage{booktabs}
\usepackage{mathrsfs}
\usepackage{amssymb} 
\usepackage{url}
\usepackage{bbm}
\usepackage{xspace,paralist}
\usepackage{times,latexsym}
\usepackage{amsmath}
\usepackage{appendix}
\usepackage{comment} 
\usepackage{enumitem}
\usepackage{makecell}
\usepackage{multirow}
\usepackage{xcolor}
\usepackage{cleveref}
\usepackage{tcolorbox}
\usepackage{adjustbox}
\usepackage{relsize}
\usepackage{todonotes}
\usepackage{natbib}
\usepackage{tabularx}

\newcommand{\appref}[1]{Appendix~\ref{#1}\xspace}

\usepackage{amssymb}% http://ctan.org/pkg/amssymb
\usepackage{pifont}% http://ctan.org/pkg/pifont

\usepackage[switch]{lineno}

\title{How Does Prefix Matter in Reasoning Model Tuning?}

\author{
  Raj Vardhan Tomar\textsuperscript{1,2}, \quad
  Preslav Nakov\textsuperscript{1}, \quad
  Yuxia Wang\textsuperscript{1,3} \\
  \textsuperscript{1}Mohamed bin Zayed University of Artificial Intelligence (MBZUAI), UAE \\
  \textsuperscript{2}Indian Institute of Technology Delhi, India 
  \textsuperscript{3}INSAIT \\
  \vspace{1.5mm}
  \texttt{\{raj.tomar, yuxia.wang\}@mbzuai.ac.ae}
}

\begin{document}
\maketitle
\begin{abstract}
Recent alignment studies commonly remove introductory boilerplate phrases from supervised fine-tuning (SFT) datasets. This work challenges that assumption. We hypothesize that safety- and reasoning-oriented prefix sentences serve as lightweight alignment signals that can guide model decoding toward safer and more coherent responses. To examine this, we fine-tune three R1 series models across three core model capabilities: reasoning (mathematics, coding), safety, and factuality, systematically varying prefix inclusion from 0\% to 100\%.

Results show that prefix-conditioned SFT improves both safety and reasoning performance, yielding up to +6\% higher Safe@1 accuracy on adversarial benchmarks (WildJailbreak, StrongReject) and +7\% improvement on GSM8K reasoning. However, factuality and coding tasks show marginal or negative effects, indicating that prefix-induced narrowing of the search space benefits structured reasoning. Token-level loss analysis further reveals that prefix tokens such as “revised” and “logically” incur higher gradient magnitudes, acting as alignment anchors that stabilize reasoning trajectories. Our findings suggest that prefix conditioning offers a scalable and interpretable mechanism for improving reasoning safety, serving as an implicit form of alignment that complements traditional reward-based methods.
% We release our dataset and code at \url{https://github.com/mbzuai-nlp/UnsafeChain}.
\end{abstract}

\section{Introduction}

Large reasoning models (LRMs) are fine-tuned on domain-specific datasets to achieve better performance on specialized tasks. Recent studies show that fine-tuning pretrained reasoning models on mathematics, programming, and scientific corpora substantially improves accuracy across GSM8K, MBPP, and GPQA benchmarks ~\cite{deepseekai2025deepseekr1incentivizingreasoningcapability, mou2025saro, zhang2025realsafe}. Beyond domain reasoning, fine-tuning remains a key mechanism for safety alignment ~\cite{tomar2025unsafechain, wang2025star, jiang2025safechain}, showing that targeted supervision can enhance robustness while maintaining general capabilities.

In safety alignment, datasets such as SafeChain~\cite{jiang2025safechain}, STAR-1~\cite{wang2025star}, and UnsafeChain~\cite{tomar2025unsafechain} curate or rewrite harmful prompts to retain safe, policy-compliant responses, yielding consistent improvements in Safe@1 accuracy on adversarial benchmarks including WildJailbreak, StrongReject, and JailbreakBench. Similarly, fine-tuning on mathematics datasets such as GSM8K~\cite{cobbe2021gsm8k,lightman2023verify} improves reasoning accuracy, while coding benchmarks such as MBPP and HumanEval~\cite{austin2021mbpp,chen2021codex} enhance program synthesis performance. Beyond reasoning and safety, recent work has extended fine-tuning to additional capabilities such as factuality~\cite{zhang2025realsafe}, demonstrating the broad applicability of domain adaptation across diverse evaluation settings.

A critical but often under-examined step in these pipelines is dataset cleaning. Most alignment datasets are preprocessed to remove prefatory boilerplate phrases introduced during prompting, such as: \textit{``Certainly! Here's a revised, well-reasoned, clear logical steps, and step-by-step explanation of the response.''} Although such prefixes superficially signal safe behavior, they do not contribute to substantive reasoning and may bias models toward hedging rather than true safety. 
By stripping these artifacts, cleaned datasets encourage models to internalize safety and correctness principles instead of imitating surface-level stylistics.

\begin{table*}[t]
    \centering
    \small
    \begin{tabular}{llll}
        \toprule
        \textbf{Task} & \textbf{Study} & \textbf{Dataset Size} & \textbf{Prompt Source} \\
        \midrule
        Safety & UnsafeChain~\cite{tomar2025unsafechain} & 13.6K & Harmful prompts from 6 datasets \\
        Safety & SafeChain~\cite{jiang2025safechain} & 40K & WildJailbreak \\
        Safety & STAR-1~\cite{wang2025star} & 1K & Harmful prompts from 18 datasets \\
        \midrule
        Mathematics & BackMATH~\cite{zhang2025backmathdata} & 14K problems & GSM8K, MATH \\
        Mathematics & Pensez~\cite{ha2025pensez} & 2K & Bilingual math problems \\
        Mathematics & SG-FT~\cite{bi2025solutionguidance} & 5K & Multi-step reasoning corpora \\
        \midrule
        Coding & MBPP-Clean~\cite{austin2021mbpp} & 1K & MBPP, HumanEval \\
        Coding & PDC \& DM-SFT~\cite{duan2025pdc} & Progressive & SQL bug-fix datasets \\
        Coding & Data-efficient CodeGen~\cite{lv2025dataefficientcode} & Unspecified & Code benchmarks \\
        \midrule
        Factuality & RealSafe-R1~\cite{zhang2025realsafe} & 15K & PKU-SafeRLHF, JailbreakV-28k \\
        Factuality & Deconv-PEFT~\cite{zhang2025peft} & 20K & General QA \\
        Factuality & DragFT~\cite{zheng2024dragft} & 30K & Domain-specific MT corpora \\
        \midrule
        Education & GuideLM~\cite{ross2025guidelm} & 528 & Student programming Qs \\
        Education & Pensez~\cite{ha2025pensez} & 2K & French math problems \\
        Education & BarExamSFT~\cite{fernandes2025barefficient} & 1,514 & MBE questions \\
        \midrule
        Healthcare & MedBioLM~\cite{kim2025medbiolm} & 50K & Biomedical QA \\
        Healthcare & VietHealth-LLM~\cite{BUI2025viethealth} & 337K & Medical forums, textbooks \\
        Healthcare & Discharge-SFT~\cite{jung2025clinical} & 17,319 & Clinical discharge notes \\
        \bottomrule
    \end{tabular}
    \caption{\textbf{Recent studies} employing dataset cleaning for LRM finetuning across tasks. Cleaning typically removes prefatory or boilerplate phrases before the substantive reasoning content. Our work revisits this assumption, asking whether eliminating these prefixes is truly necessary for improved alignment and reasoning quality. See study-wise cleaning method in Table~\ref{tab:previous-study-cleaning-methods} in Appendix~\ref{app:previous-study=cleaning-method}.}
    \label{tab:cleaningstudies}
\end{table*}

Table~\ref{tab:cleaningstudies} summarizes representative studies across safety, mathematics, coding, factuality, education, and healthcare that adopt this cleaning strategy. While removing prefatory content is widely regarded as best practice, recent evidence suggests that small, structured linguistic cues can meaningfully influence model behavior. LookAhead~\cite{liu2025lookahead} demonstrates that short prefix previews can guide decoding toward safer trajectories.

Motivated by these findings, we revisit a fundamental question: is eliminating prefatory prefixes during SFT always desirable? We hypothesize that selective prefix retention may instead serve as a lightweight alignment mechanism, shaping optimization dynamics and decoding behavior in ways that benefit reasoning and safety. To investigate this, we conduct a systematic study of prefix inclusion during SFT across three core model capabilities: reasoning (mathematics and coding), safety, and factuality, while holding all other training and inference conditions fixed. We further compute the average per-token cross-entropy loss during SFT to see if these prefixes are \textit{harder to predict} and thus produce larger gradient updates.

Our experiments confirm that prefix retention substantially benefits reasoning heavy and safety critical domains.  
The results reveal that lightweight prefix sentences, even a single guiding phrase, can improve reasoning coherence and safety behavior. This work makes the following key contributions:
\begin{itemize}
  \setlength{\itemsep}{0pt}
  \setlength{\parskip}{0pt}
  \setlength{\parsep}{0pt}
  \item A principled synthesis linking prefix-level interventions and token-level training dynamics (building on prefix-tuning and token-sensitivity literature).
  \item A controlled SFT ablation that isolates prefix prevalence as the sole manipulated variable across capabilities and domains.
  \item Empirical analysis shows that prefix tokens consistently attract higher per-token loss, and that limited prefix exposure improves reasoning and safety performance while yielding mixed effects on coding and factuality, providing a loss-based explanation for how small, semantically meaningful prefixes function as efficient alignment scaffolds during SFT.
\end{itemize}
\section{Related Work}

Recent efforts to align large reasoning models (LRMs) with human preferences and safety objectives have largely converged into three methodological directions: supervised fine-tuning (SFT) and reinforcement learning (RL) for safety alignment, inference-time scaling and decoding modifications, and external guard models. Our work is most closely related to the first line, with a particular focus on the role of dataset preprocessing and prefix manipulation in safety alignment.

\paragraph{Dataset Cleaning in Domain-Specific Finetuning}
Beyond safety specific datasets, domain finetuning commonly relies on dataset cleaning to improve coherence and accuracy. In mathematics, BackMATH~\cite{zhang2025backmathdata} and Pensez~\cite{ha2025pensez} remove templated boilerplate and spurious reasoning steps, while SolutionGuidance FT (SG-FT)~\cite{bi2025solutionguidance} standardizes reasoning structure through auxiliary guidance. In coding, MBPP~\cite{austin2021mbpp}, PDC~\cite{duan2025pdc}, and DragFT~\cite{zheng2024dragft} strip template phrases and noisy artifacts while incorporating retrieval- or dictionary-based augmentation. For factuality, Deconv-PEFT~\cite{zhang2025peft} and RealSafe-R1~\cite{zhang2025realsafe} emphasize removing hallucinated or self-referential content and enforcing explicit refusals. Education and healthcare datasets, GuideLM~\cite{ross2025guidelm}, BarExamSFT~\cite{fernandes2025barefficient}, VietHealth-LLM~\cite{nguyen2025viethealth}, and Discharge-SFT~\cite{jung2025clinical}, similarly apply domain-specific cleaning heuristics to ensure concise and trustworthy outputs.

\paragraph{Every Token Matters}
A growing body of work shows that large reasoning models (LRMs) are highly sensitive to small, localized interventions at both the token and parameter levels. Alignment-Enhanced Decoding (AED)~\cite{liu2024alignmentenhanced} demonstrates that safety-helpfulness conflicts often emerge at early decoding steps, and that reweighting a small set of tokens can substantially reduce jailbreak success. Similarly, Token Highlighter~\cite{hu2024tokenhighlighter} identifies a few affirmation tokens (e.g., ``Sure, here is...'') that contribute to unsafe behavior, showing that softly suppressing them thwarts multiple jailbreak families. Prompt-based attacks~\cite{mustafa2025anyonejailbreak} and token-mining methods such as JAILMINE~\cite{li2024lookpicking} further illustrate that inserting or mutating only a handful of tokens can bypass alignment defenses. At the parameter level, super-weight analyses~\cite{yu2024superweight} reveal that individual scalar weights can dominate model behavior, suggesting a broader principle: just as a single parameter can determine model fidelity, a small number of strategically placed tokens, whether added, removed, or reweighted, can alter safety and generalization. Together, these findings motivate closer scrutiny of data cleaning and prefix engineering in LRM alignment.

\section{Problem Definition}

\subsection{Motivation}
%  and synthesis of prior work
\paragraph{Prefix Tuning}
Recent work has emphasized the critical role of prefixes in steering LLMs. Traditional prefix tuning ~\cite{li2021prefixtuning} introduced the idea of prepending continuous, learnable “virtual tokens” to each transformer layer, enabling parameter-efficient adaptation. Prefix-Tuning+ ~\cite{wang2025prefixtuningplus} identified that standard prefix tuning suffers from a trade-off between prefix significance and input fidelity when embedded within attention heads. To overcome this, they proposed relocating the prefix module outside the attention mechanism, reformulating it as an external learnable matrix.

Unsupervised Prefix Fine-Tuning (UPFT) ~\cite{ji2025upft} used the observation of prefix self-consistency, the phenomenon that early reasoning steps across multiple trajectories are highly similar. By fine-tuning only on the first few tokens of generated responses (as few as 8 tokens), UPFT achieved reasoning improvements comparable to supervised rejection-sampling fine-tuning.

From a post-training perspective, Prefix-RFT ~\cite{huang2025prefixrft} proposed integrating supervised fine-tuning (SFT) and reinforcement fine-tuning (RFT) by sampling prefixes from demonstrations and mixing them with on-policy rollouts. The prefix serves as a guiding signal during reinforcement updates, combining the stability of demonstrations with the flexibility of exploration.

\paragraph{Prefixes as Lightweight Alignment Signals.}
While cleaning has become standard practice, recent studies challenge the assumption that prefixes are purely detrimental. LookAhead Tuning~\cite{liu2025lookahead} demonstrated that introducing short prefix previews before the main answer can improve model robustness by constraining decoding trajectories, enabling models to anticipate unsafe continuations. This aligns with the “less is more” (LIMO) philosophy in LRM alignment~\cite{ye2025limo}, which emphasizes that even minimal amounts of carefully chosen data or guiding tokens can produce disproportionately large improvements in generalization and reasoning performance. In our setting, prefix tokens function as lightweight alignment signals that bias early decoding trajectories, effectively narrowing the model’s search space with minimal additional supervision.

Together, these findings suggest prefixes as candidate \emph{alignment signals}: short, semantically meaningful token sequences that (i) occur in early decoding positions, (ii) may be harder to predict during SFT, and (iii) therefore can concentrate gradients and bias subsequent generation. This frames prefixes as lightweight alignment cues, motivating a systematic re-examination of their role in dataset construction and model finetuning.

\subsection{Research Questions}
Building on these observations, we investigate the following questions:
\begin{itemize}
    \setlength{\itemsep}{0pt}
    \setlength{\parskip}{0pt}
    \setlength{\parsep}{0pt}
    \item \textbf{RQ1:} Does retaining prefatory prefixes during supervised fine-tuning affect reasoning, safety, or factuality performance?
    \item \textbf{RQ2:} Are the effects of prefix retention consistent across different model capabilities and application domains?
    \item \textbf{RQ3:} Can token-level training dynamics explain why a small number of prefix tokens influence downstream behavior?
\end{itemize}

\subsection{Hypothesis}
We hypothesize that prefix tokens function as high-impact alignment signals during SFT. Because these tokens appear at the beginning of responses and are semantically loaded, they are harder to predict and thus incur higher training loss. Under cross-entropy optimization, this leads to larger gradient updates, allowing a small number of prefix tokens to disproportionately influence model updates and subsequent decoding behavior. By shaping gradient flow during SFT, prefixes act as implicit “policy reminders,” narrowing the model's search space through supervised fine-tuning (SFT) before reinforcement learning (RL), thereby allowing RL to rely on simpler reward models for refinement. This view reframes dataset preprocessing not just as noise removal, but as a mechanism for injecting controlled inductive bias into reasoning models.

\section{Experiments}

\subsection{Models and SFT Setup}
We experiment with three unaligned distilled reasoning models from \citet{deepseekai2025deepseekr1incentivizingreasoningcapability}.
They are \texttt{DeepSeek-R1-Distill-Qwen-1.5B/7B} and \texttt{DeepSeek-R1-Distill-Llama-8B}.
They were selected since \citet{zhou2025hiddenrisk} showed that distilled reasoning models exhibit the weakest safety and reasoning performance compared to both closed- and open-source models trained to reason via direct Reinforcement Learning.

We fine-tune these models on GSM8K, UnsafeChain, MBPP, and TruthfulQA using Parameter-Efficient Fine-Tuning (PEFT) with LoRA adapters ~\cite{hu2022lora}. All models were trained for 2 epochs using mixed-precision FP16 on a single NVIDIA A6000 GPU, with 8-bit model loading and gradient accumulation. LoRA is applied to the \texttt{q\_proj} and \texttt{v\_proj} modules. This uniform setup ensures controlled comparisons across datasets. See \appref{appendix:finetune-config} for full training configuration. During training, we log the per-token loss and gradient norms at each step to quantify how prefix tokens influence convergence and internal representation stability.

\subsection{Datasets}

\begin{table*}[t!]
    \centering
    % \small
    \resizebox{0.95\textwidth}{!}{
    \begin{tabular}{lll}
        \toprule
        \textbf{Tasks} & \textbf{Training Data} & \textbf{Test Data} \\
        \midrule
        Safety &  1K UnsafeChain & 200 each from WildJailbreak, StrongReject, WildChat, JailbreakBench  \\
        Mathematics & 1K GSM8K & 300 GSM8K test, 300 Math500 \\
        Coding & 774 MBPP & 200 MBPP, 164 HumanEval \\
        Factuality & 617 TruthfulQA & 200 each from TruthfulQA (TQA) and TruthfulQA-MCQ \\
        \bottomrule
    \end{tabular}
    }
    \caption{\textbf{Datasets for training and evaluation.} Training samples were rewritten using GPT-4.1 with reasoning style. Prefix sentences were reintroduced in varying proportions.} %  JBB: JailBreak Bench, WJ: Wild Jailbreak, SR: Strong Reject, and TQA: TruthfulQA
    \label{tab:datasets}
\end{table*}

We evaluate three core model capabilities: reasoning (mathematics and coding), safety, and factuality.
For each capability, we begin with a base dataset
\[
D = \{(x_i, y_i)\}_{i=1}^{N},
\]
where \(x_i\) denotes the original task prompt (e.g., a math problem, or factual question), and \(y_i\) is the corresponding target response rewritten using GPT-4.1 to follow a standardized reasoning format.
Each rewritten response consists of step-by-step reasoning enclosed in \texttt{<think>} tags followed by a final answer.
During this rewriting stage, the model is instructed \emph{not} to include any prefatory boilerplate or safety-style prefixes. For each capability, we define a fixed set of five prefix templates
\[
S = \{s_1, s_2, s_3, s_4, s_5\},
\]
designed to reflect common stylistic scaffolds observed in instruction-tuning data.
Concrete examples of these prefixes are provided in Appendix~\ref{appendix:prefix-examples}.

Given a prefix inclusion ratio \(\alpha \in \{0, 0.25, 0.5, 1.0\}\), we construct a prefix-augmented training set \(D_{\alpha}\) as follows.
For each training example \((x_i, y_i)\), a prefix is added independently with probability $\alpha$.
If \(z_i = 1\), we uniformly sample a prefix \(s_j \sim \mathrm{Uniform}(S)\) and form the prefixed input
\[
\tilde{x}_i = s_j \,\Vert\, x_i,
\]
where \(\Vert\) denotes string concatenation.
If \(z_i = 0\), we set \(\tilde{x}_i = x_i\).
The resulting dataset is
\[
D_{\alpha} = \{(\tilde{x}_i, y_i)\}_{i=1}^{N}.
\]

This construction ensures that (i) the underlying task distribution remains unchanged across conditions, (ii) each prefix template is equally likely when prefixes are applied, and (iii) prefix prevalence is the only controlled variable across runs.

For mathematics reasoning, we further introduce a \emph{keyword augmented} variant of the prefix set, in which each prefix \(s_j\) is minimally modified to include the single keyword \textit{``revised''}.
This results in two prefix families, original and revised, allowing us to isolate the effect of a single high-signal token while holding all other factors constant.
Examples of the original and revised prefix pairs are provided in Appendix~\ref{appendix:prefix-examples}. Table~\ref{tab:datasets} summarizes the training and evaluation datasets used for each capability.

\begin{table*}[t!]
    \centering
    \small
    % \resizebox{\textwidth}{!}{
    \begin{tabular}{lccccc}
        \toprule
        \textbf{Model} & \textbf{Prefix (\%)} & \textbf{GSM8K} & \textbf{GSM8K-Rev} & \textbf{Math500} & \textbf{Math500-Rev} \\
        \midrule
        \multirow{4}{*}{\textbf{R1-8B}} & 0  & 65.67  & -- & \textbf{70.67}  & -- \\
        & 25  & 69.33  & 71.00  & 67.33  & 65.67  \\
        & 50  & 70.00  & 67.00  & 67.67  & 69.33  \\
        & 100  & \textbf{72.33}  & 70.33  & 69.67  & 69.67  \\
        \midrule
        \multirow{4}{*}{\textbf{R1-7B}} & 0  & 87.67  & -- & 78.33  & -- \\
        & 25  & \textbf{89.00}  & 84.00  & 75.00  & 76.33  \\
        & 50  & 88.33  & 89.00  & 75.33  & 78.33  \\
        & 100  & 88.00  & 85.00  & \textbf{78.67}  & 78.67  \\
        \midrule
        \multirow{4}{*}{\textbf{R1-1.5B}} & 0  & \textbf{55.00}  & -- & 48.67  & -- \\
        & 25  & 52.67  & 54.00  & \textbf{50.33}  & 51.00  \\
        & 50  & \textbf{55.00}  & 54.33  & 48.00  & 47.67  \\
        & 100  & 51.67  & 52.00  & 46.33  & 46.67  \\
        \bottomrule
    \end{tabular}
    % }
    \caption{\textbf{Reasoning Mathematics Results} of models finetuned on 1,000 GSM8K training examples (\textbf{GSM8K 1K}). under different prefix inclusion ratios (0\%, 25\%, 50\%, and 100\%). Each prefix ratio corresponds to a separate supervised fine-tuning run using the same training set. All models are evaluated on held-out test sets, specifically the GSM8K test split and Math500, which are never seen during training. The ``Revised Variant'' (-Rev) column reports runs where the prefix scaffold explicitly contained the word \textit{``revised''} within the opening phrase, reflecting a single-token alignment cue. Best results in \textbf{bold}.}
    \label{tab:maths}
\end{table*}

\subsection{Evaluation Setup}

We evaluated model behavior across three core model capabilities: reasoning (mathematics, coding), safety, and factuality, under multiple decoding strategies. 
Following the UnsafeChain protocol~\cite{tomar2025unsafechain}, we used GPT-4.1 as the primary judge for scoring correctness, and quality unless otherwise specified. 
For safety classification, when not scored by GPT-4.1, we used \texttt{Llama-Guard-3-8B}~\cite{dubey2024llama3herdmodels} to determine whether a response is safe. We also log per-token loss across test examples to analyze the alignment between prefix tokens and token-level gradient influence. All evaluations are performed under greedy decoding ($T=1.0$) to minimize randomness and ensure consistent comparison across prefix conditions. Detailed descriptions of each evaluation benchmark and metric are in Appendix~\ref{appendix:evaluation}.

\section{Results}

We systematically evaluate the effect of prefix inclusion across safety, reasoning, coding, and factuality domains.
The results, summarized in Tables ~\ref{tab:maths}, \ref{tab:safety}, \ref{tab:coding}, and \ref{tab:factuality}, reveal that structured prefix sentences meaningfully alter model behavior, influencing both reasoning accuracy and safety alignment.
Prefix-guided supervision improves reasoning consistency and safe refusal behaviors but can sometimes constrain factual precision and open-ended generation.
Our experiments also show that common prefix tokens consistently have much higher training loss than average tokens, as shown in Table~\ref{tab:loss}. Because higher loss corresponds to larger gradient updates under cross-entropy training, a small number of prefix tokens can disproportionately influence model optimization.

\begin{figure}[t!]
    \centering
    \includegraphics[width=0.85\linewidth]{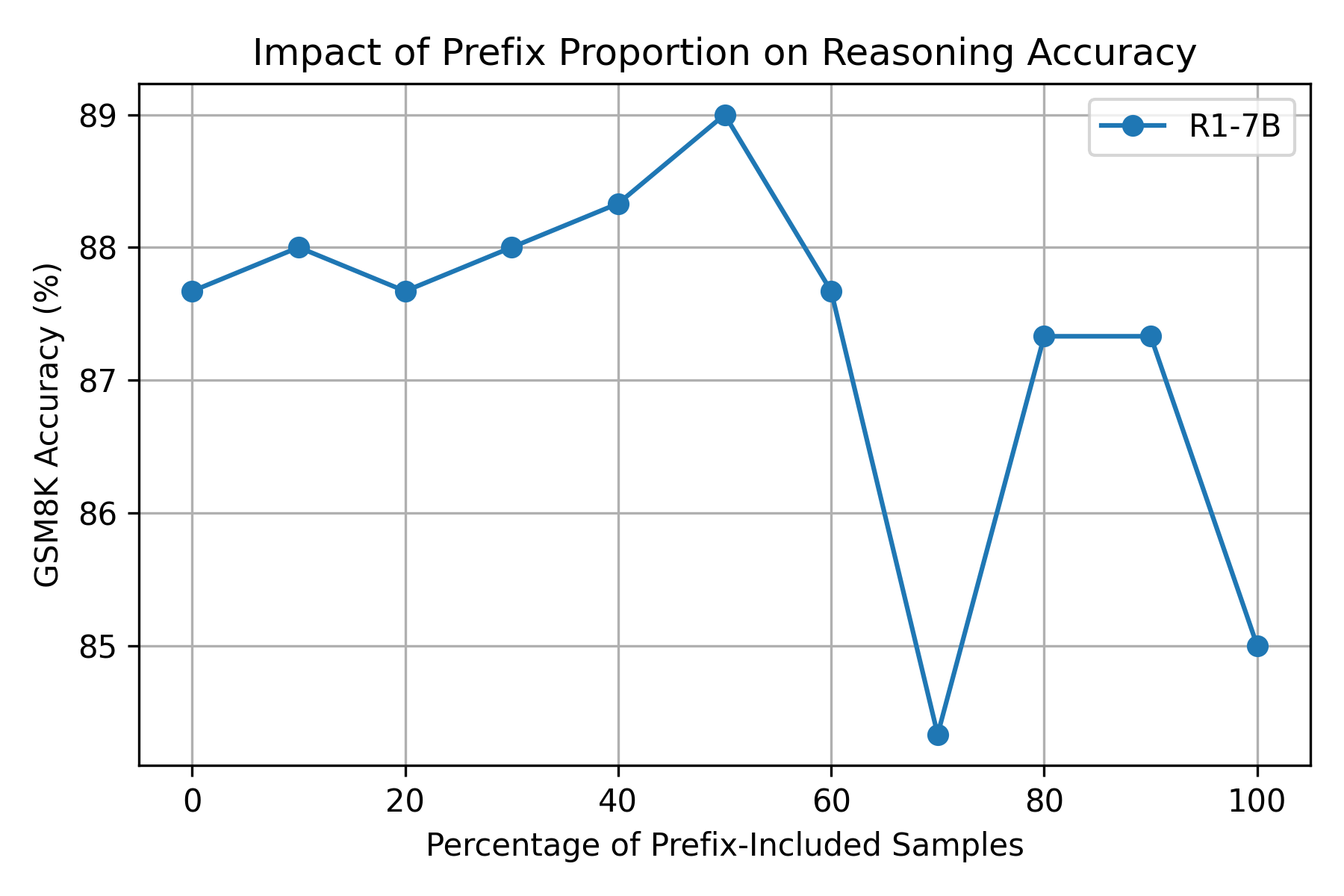}
    \caption{\textbf{Impact of prefix inclusion on reasoning performance.} 
    GSM8K accuracy (\%) for the R1-7B model trained with increasing proportions of prefix-included samples with "revised" token added to the prefix.}
    \label{fig:gsm8k_curve}
\end{figure}

\paragraph{Math}
Table~\ref{tab:maths} presents results on GSM8K and Math500.
Reasoning accuracy improves for models fine-tuned with prefix scaffolds.
The addition of the “revised” keyword within the prefix scaffold consistently stabilizes intermediate reasoning steps, suggesting that single-token alignment cues can reinforce task framing during decoding.
Figure~\ref{fig:gsm8k_curve} further illustrates this behavior: GSM8K accuracy peaks at intermediate prefix ratios, forming an inverted-V accuracy curve.
This indicates that limited exposure to guiding prefixes improves reasoning structure by reducing ambiguity in the model’s initial decoding trajectory, whereas excessive prefix repetition introduces redundancy and degrades calibration.
Prefix-guided finetuning achieves a similar effect by implicitly encoding “reasoning priors” in the opening tokens, allowing models to converge faster to semantically consistent solutions.

\begin{table}[t!]
    \centering
    \resizebox{\columnwidth}{!}{
    \begin{tabular}{lccccc}
        \toprule
        \textbf{Model} & \textbf{Prefix (\%)} & \textbf{WJ} & \textbf{SR} & \textbf{WC} & \textbf{JBB} \\
        \midrule
        \multirow{4}{*}{\textbf{R1-8B}} & 0  & 60.00  & 36.00  & \textbf{96.00}  & 53.33  \\
        & 25  & 62.80  & 35.00  & 94.50  & 58.67  \\
        & 50  & \textbf{66.00}  & 38.00  & 95.50  & 60.00  \\
        & 100  & 64.40  & \textbf{41.00}  & 95.00  & \textbf{61.00}  \\
        \midrule
        \multirow{4}{*}{\textbf{R1-7B}} & 0  & 60.40  & 23.00  & 96.00  & 50.33  \\
        & 25  & 61.20  & 25.00  & \textbf{96.50}  & 54.00  \\
        & 50  & \textbf{64.00}  & \textbf{30.00}  & 94.00  & 53.67  \\
        & 100  & 62.40  & 26.00  & 95.00  & \textbf{55.33}  \\
        \midrule
        \multirow{4}{*}{\textbf{R1-1.5B}} & 0  & \textbf{40.00}  & 6.00  & 93.50  & 44.67  \\
        & 25  & 38.00  & \textbf{7.00}  & \textbf{96.50}  & \textbf{48.67}  \\
        & 50  & 39.60  & \textbf{7.00}  & 95.50  & 47.67  \\
        & 100  & \textbf{40.00}  & 5.00  & 94.00  & \textbf{48.67}  \\
        \bottomrule
    \end{tabular}
    }
    \caption{\textbf{Safety Results} of models fine-tuned on 1,000 UnsafeChain training samples (\textbf{UnsafeChain 1K}) under varying prefix inclusion ratios (0\%, 25\%, 50\%, 100\%) and Safe@1 is calculated on held-out safety benchmarks: WildJailbreak (WJ), StrongReject (SR), WildChat (WC), and JailbreakBench (JBB). Each prefix ratio corresponds to an independently fine-tuned model. Best results in \textbf{bold}.}
    \label{tab:safety}
\end{table}

\begin{table}[t!]
\centering
\small
% \begin{minipage}[t]{0.48\textwidth}
% \centering
% \resizebox{\linewidth}{!}{
\begin{tabular}{lccc}
\toprule
\textbf{Model} & \textbf{Prefix (\%)} & \textbf{MBPP} & \textbf{HumanEval} \\
\midrule
\textbf{R1-8B} & 0  & 49.50 & \textbf{64.02} \\
& 25  & 48.50 & 57.93 \\
& 50  & 48.50 & 59.76 \\
& 100  & \textbf{50.00} & 60.98 \\
\midrule
\textbf{R1-7B} & 0  & \textbf{45.00} & 70.73 \\
& 25  & 41.50 & 67.07 \\
& 50  & 41.00 & 69.51 \\
& 100  & 37.50 & \textbf{71.34} \\
\midrule
\textbf{R1-1.5B} & 0  & \textbf{31.00} & \textbf{17.68} \\
& 25  & 28.00 & 15.24 \\
& 50  & 25.50 & 15.24 \\
& 100  & 30.00 & 17.07 \\
\bottomrule
\end{tabular}
% }
\caption{\textbf{Reasoning Coding Results} of models fine-tuned on 1000 MBPP training examples (\textbf{MBPP 1K}) under varying prefix inclusion ratios (0\%, 25\%, 50\%, 100\%) and evaluated on held-out MBPP and HumanEval benchmarks. Each ratio corresponds to an independently fine-tuned model. Best results in \textbf{bold}.}
\label{tab:coding}
% \end{minipage}
% \hfill
\end{table}

\begin{table}
% \begin{minipage}[t]{0.48\textwidth}
\centering
\small
% \resizebox{\linewidth}{!}{
\begin{tabular}{lccc}
\toprule
\textbf{Model} & \textbf{Prefix (\%)} & \textbf{TQA} & \textbf{TQA-MCQ} \\
\midrule
\textbf{R1-8B} & 0  & \textbf{39.50} & \textbf{56.50} \\
& 25  & 39.00 & 48.50 \\
& 50  & 38.00 & 53.00 \\
& 100  & 36.00 & 47.50 \\
\midrule
\textbf{R1-7B} & 0  & 20.00 & \textbf{43.50} \\
& 25  & 20.50 & 41.50 \\
& 50  & \textbf{24.00} & 37.50 \\
& 100  & 23.50 & 39.00 \\
\midrule
\textbf{R1-1.5B} & 0  & \textbf{5.00} & \textbf{31.50} \\
& 25  & 3.50 & 27.50 \\
& 50  & 3.50 & 28.00 \\
& 100  & 2.50 & 23.50 \\
\bottomrule
\end{tabular}
% }
\caption{\textbf{Factuality Results} of models fine-tuned on 1,000 TruthfulQA training samples (\textbf{TQA 1K}) under varying prefix inclusion ratios (0\%, 25\%, 50\%, 100\%) and evaluated on held-out TruthfulQA and TruthfulQA-MCQ benchmarks. Each prefix ratio corresponds to an independently fine-tuned model. Best results in \textbf{bold}.}
\label{tab:factuality}
% \end{minipage}
\end{table}
\paragraph{Safety}
Table~\ref{tab:safety} reports Safe@1 results on adversarial benchmarks under varying prefix inclusion ratios.
Prefix conditioning yields consistent safety improvements across all scales, with the most pronounced effects on WildJailbreak and StrongReject.
These results confirm that short alignment-prefixed statements act as semantic initialization priors, biasing model trajectories toward refusal-consistent regions of the decoding space.
The persistence of this effect across models suggests that prefixes act as low-dimensional control signals, influencing early contextual embedding states to enforce safety constraints without explicit reward modeling.

\paragraph{Coding and Factuality}
For coding in Table~\ref{tab:coding}, MBPP and HumanEval accuracies show weak or inconsistent correlations with prefix inclusion.
This pattern suggests that prefix conditioning primarily aids reasoning-style generation rather than syntax-driven completion.

In factuality tasks (TruthfulQA and TruthfulQA-MCQ) in Table~\ref{tab:factuality}, prefix inclusion consistently reduces accuracy across all model sizes.
Unlike reasoning-oriented benchmarks, TruthfulQA performance primarily depends on the breadth and reliability of the model’s pretraining data, rather than on stylistic coherence or stepwise reasoning.
The addition of alignment-oriented prefixes appears to constrain the model’s factual recall, steering responses toward normative or interpretive phrasing instead of evidence-grounded answers.
Thus, while prefix scaffolds enhance structured reasoning and safety behaviors, they may inadvertently suppress factual precision in knowledge-intensive domains where accurate information retrieval is the key determinant of success.

\paragraph{Why is Prefix Important?}

\begin{table*}[t!]
    \centering
    \small
    % \resizebox{0.9\linewidth}{!}{
    \begin{tabular}{lcccccc}
        \toprule
        \textbf{Domain} & \textbf{All Tokens} & \textbf{Certainly} & \textbf{Revised} & \textbf{Rewritten} & \textbf{Logically} & \textbf{Response} \\
        \midrule
        Safety & 3.58 & 9.18 & 11.10 & 10.96 & 10.55 & 5.05 \\
        Maths & 2.07 & 12.60 & 11.22 & 13.28 & 10.96 & 10.26 \\
        Factuality & 2.97 & 9.52 & 9.47 & 10.27 & 11.34 & 4.41 \\
        Coding & 1.80 & 10.82 & 7.82 & 6.69 & 11.38 & 3.32 \\
        \bottomrule
    \end{tabular}
    % }
    \caption{\textbf{Token-level loss concentration across domains.} Average per-token loss values computed on 100\% prefix-included samples for the R1-series models.}
    \label{tab:loss}
\end{table*}

Recent work has shown that analyzing training dynamics at the level of individual tokens provides valuable insight into model behavior and alignment. Rather than treating the loss as a single global scalar, token-level analyses reveal how specific words, prefixes, or reasoning cues influence gradient flow and representation updates. ToDi~\cite{jung2025todi} computes divergence losses at the token level to better control alignment with teacher models, while SOC~\cite{zhang2025soc} frames supervised fine-tuning as a sequence of token-level decisions and introduces auxiliary losses to correct overconfident predictions. Related work on vocabulary frequency imbalance~\cite{chung2025vocabimbalance} further shows that loss reduction concentrated on a subset of tokens can account for a large fraction of training gains. Together, these studies motivate using per-token loss as an interpretable diagnostic for understanding how models internalize alignment signals.

Inspired by these findings, we examine prefix influence through token-level training loss. Specifically, for each training run, we record the cross-entropy loss at every token position during supervised fine-tuning. We compute the average loss for frequently occurring prefix tokens (\textit{``Certainly''}, \textit{``revised''}, \textit{``response''}, \textit{``logically''}) by aggregating their losses across all occurrences in prefix-included samples and averaging over training steps. As a reference, we also compute the corpus-wide average loss across all tokens within the same runs. The resulting values are reported in Table~\ref{tab:loss}.

Across all evaluated settings, prefix tokens consistently exhibit substantially higher average loss (typically 9--13) than the corpus-average token loss (2--3). Under standard cross-entropy optimization, higher loss corresponds to larger gradient magnitudes, indicating that these tokens receive disproportionately strong parameter updates during training. This effect is most pronounced in mathematics datasets, where the loss for \textit{``Certainly''} reaches 12.6 compared to an overall average of 2.07. While our experiments do not directly probe layer-wise representations, the observed loss concentration supports the interpretation consistent with prior token-level analyses~\cite{hu2024tokenhighlighter} that a small number of semantically salient prefix tokens can dominate gradient flow and thereby influence downstream generation behavior.

Model scale significantly mediates prefix sensitivity.  
Larger reasoning models like R1-7B and R1-8B, demonstrated strong robustness to prefix inclusion, maintaining or improving performance under noisy or full-prefix conditions.  
In contrast, R1-1.5B exhibited brittle behavior, showing fluctuations in factual and coding benchmarks.  
This scale dependent trend indicates that larger models are capable of interpreting prefix tokens as contextual alignment cues, while smaller models overfit to them and treat prefixes as literal input content.  

Overall, these results suggest that prefix-guided SFT functions as a lightweight alignment mechanism: a small set of high-loss prefix tokens can inject strong alignment signal during training, leading to measurable changes in reasoning and safety behavior without requiring large-scale data augmentation or explicit reward modeling.
\section{Discussion}

Our study shows that selectively retaining prefatory prefixes during supervised fine-tuning can meaningfully alter the behavior of large reasoning models, but that these effects are strongly capability dependent. 
Our findings reveal broader implications for the alignment of reasoning-capable models.

\paragraph{Why do prefixes help safety and mathematical reasoning but hurt coding and factuality?}
One pattern across our experiments is that prefix inclusion tends to improve performance on safety and mathematical reasoning, while yielding weaker or negative effects on coding and factuality. This divergence arises from differences in task structure and the role of intermediate reasoning.

Safety and mathematical reasoning tasks benefit from structured, stepwise generation and from early commitment to a normative or policy-aligned mode of response. In these settings, prefatory prefixes may act as soft constraints that bias the model toward refusal-consistent or logically organized continuations, reducing ambiguity in early decoding and encouraging coherent reasoning trajectories. This interpretation is consistent with the observed gains on adversarial safety benchmarks and with the non-monotonic accuracy trends on GSM8K.

In contrast, coding and factuality tasks place greater emphasis on precision and direct content retrieval. Coding benchmarks require syntactically exact outputs with minimal natural-language scaffolding, and factuality benchmarks reward accurate recall over explanatory style. In these regimes, prefix-induced biases toward verbosity, normative framing, or interpretive language may interfere with the model’s ability to emit concise code or retrieve specific facts, thereby degrading performance. This suggests that prefix scaffolds introduce an inductive bias that is beneficial for tasks requiring structured reasoning or refusals, but misaligned with tasks dominated by exactness.

\paragraph{Mechanistic Interpretation.}
Integrating the evidence from Tables~\ref{tab:maths}–\ref{tab:loss} and Figure~\ref{fig:gsm8k_curve}, we can say that prefix inclusion appears to inject a low-entropy initialization signal that improves gradient flow during reasoning token generation and strengthens early-layer semantic conditioning.  
Hence, prefix retention serves as a compact, data-level intervention—effectively steering the representational dynamics of LRMs without the need for explicit reward modeling or large-scale instruction data.

\paragraph{Rethinking Dataset Cleaning and RLHF}
Earlier studies emphasized removing boilerplate sentences during dataset cleaning to avoid stylistic bias. Our results challenge this assumption by demonstrating that prefixes can serve as alignment scaffolds when incorporated strategically.  
This finding raises a question: should alignment pipelines prioritize cleaning or optimizing prefixes?  
Given their low cost and high impact, prefix based conditioning could simplify RLHF by pre-aligning model trajectories during SFT.  
If the model’s output distribution is already biased toward safe continuations, the reward model’s burden shifts from enforcing safety to refining preference granularity.  
Thus, prefix optimization offers a scalable complement to traditional RLHF embedding alignment into model representations before reinforcement fine-tuning.
\section{Conclusion and Future Work}

This work examined whether prefix sentences can systematically improve the reasoning and alignment behavior of LRMs during SFT.  
Our results show that even minimal prefix inclusion can enhance Safe@1 accuracy and mathematical reasoning quality, while its effects on factuality and coding remain task-dependent.  
Loss-level analysis revealed that prefix tokens such as \textit{“revised”} and \textit{“certainly”} carry disproportionately high per-token losses, acting as alignment anchors that receive strong gradient updates and steer the model’s early reasoning trajectory.  

Our study underscores that every token matters in the alignment of reasoning models.  
Prefix retention is not merely a stylistic artifact but a computationally meaningful signal that guides representation learning, attention allocation, and decoding dynamics.  
We reveal a new paradigm for lightweight alignment interventions, one that balances reasoning coherence, safety behavior, and interpretability within a unified framework.
This challenge the conventional assumption that boilerplate prefixes must be cleaned from instruction-tuning data, instead suggesting that small but well-chosen prefix cues can enhance reasoning and safety capabilities.  

\paragraph{Future Work}
Several directions remain open for future investigation:  
(i) evaluating prefix-conditioned fine-tuning on larger frontier models (e.g., 70B-405B) to assess scaling effects;  
(ii) studying how prefix-guided SFT interacts with downstream alignment methods like RLHF and DPO; and  
(iii) exploring automated prefix discovery through gradient based prefix optimization.

\section*{Limitations}

\paragraph{Model Scale and Scope.}
Our experiments are conducted on distilled reasoning models up to 8B parameters. While these models are appropriate for studying alignment sensitivity and prefix effects, the results may not directly generalize to larger frontier models or to models trained with full reinforcement learning. Larger models may interpret prefix cues differently or exhibit reduced sensitivity due to greater representational capacity.

\paragraph{Dependence on Automated Data Generation.}
Training data across reasoning, safety, coding, and factuality tasks are rewritten using GPT-4.1 to enforce a standardized reasoning format. Although this enables controlled comparisons, it also introduces reliance on a single strong model for data curation. As a result, stylistic or reasoning biases inherent to GPT-4.1 may be reflected in the fine-tuned models, potentially limiting diversity in reasoning styles.

\paragraph{Automated Evaluation and Judgment Noise.}
Our evaluation pipeline relies on automated judges, including GPT-4.1 for correctness and quality assessment and LlamaGuard for safety classification. While these tools provide scalability and consistency, they may fail to capture nuanced errors, borderline safety violations, or subtle factual inaccuracies. In particular, safety classifiers can mislabel complex or ambiguous cases, which may affect reported Safe@1 scores.

\section*{Ethical Statements}

This work aims to improve safety and alignment in large reasoning models by studying the role of prefix conditioning during supervised fine-tuning. While our methodology prioritizes controlled experimentation and safety-aware evaluation, several ethical considerations merit discussion.

\paragraph{Use of Adversarial and Sensitive Prompts.}
Our datasets include adversarial, policy-violating, and potentially harmful prompts drawn from established benchmarks. These prompts are used strictly for research purposes to evaluate and improve model robustness. All unsafe or harmful content is processed through controlled rewriting and safety evaluation pipelines, and models are not deployed in real-world settings based on these experiments.

\paragraph{Reliance on Automated Safety Tools.}
Safety evaluation in this work depends on LlamaGuard, an automated classifier with predefined safety policies. While such tools are standard in alignment research, they encode implicit policy assumptions and may not fully reflect diverse cultural, legal, or contextual norms. Misclassification of responses—either false positives or false negatives—remains a risk.

\paragraph{Normative Bias in Alignment Signals.}
Prefix scaffolds and safety-oriented cues reflect implicit judgments about what constitutes appropriate or aligned behavior. These judgments are informed by common safety heuristics in current alignment research but may not generalize across all deployment contexts. Users applying prefix-based alignment techniques should consider additional domain-specific and cultural evaluations.

\paragraph{No Human Subjects or Annotator Risk.}
This study does not involve human annotators or user data. All training data rewriting and evaluation are performed using automated models, eliminating direct exposure of human workers to harmful content. However, this also increases reliance on the limitations of automated judgments, which may overlook subtle ethical concerns.

Overall, we view prefix-based alignment as a lightweight and controllable intervention that complements existing safety practices, but not as a substitute for broader governance, human oversight, or deployment-time safeguards.

\bibliographystyle{acl_natbib}
\bibliography{main}

\newpage
\appendix
\clearpage
\appendix
% \section*{\centering {\LARGE \textbf{Appendix}}}
\section*{Appendix}
\addcontentsline{toc}{section}{Appendix}

\section{Fine-Tuning Configuration}
\label{appendix:finetune-config}

We fine-tuned all models using Parameter-Efficient Fine-Tuning (PEFT) with LoRA adapters:

\begin{itemize}
    \item \textbf{LoRA Config:} Rank = 16, Alpha = 32, Dropout = 0.05
    \item \textbf{Target Modules:} \texttt{q\_proj}, \texttt{v\_proj}
    \item \textbf{Training Hyperparameters:}
    \begin{itemize}
        \item Learning Rate = 1e-5
        \item Epochs = 2
        \item Gradient Accumulation = 8
    \end{itemize}
    \item \textbf{Precision/Hardware:}
    \begin{itemize}
        \item Mixed-precision (FP16), 8-bit loading
        \item NVIDIA A6000 GPU (48GB VRAM)
    \end{itemize}
\end{itemize}

\section{Evaluation Benchmarks and Metrics}
\label{appendix:evaluation}

We evaluate models across three core capabilities: reasoning, safety, and factuality, using established benchmarks. Below we summarize each dataset and its evaluation protocol.

\begin{itemize}
    \item \textbf{WildJailbreak}~\cite{wildteaming2024}: a set of adversarial jailbreak prompts designed to elicit unsafe responses. We report Safe@1 accuracy.

    \item \textbf{StrongReject}~\cite{souly2024strongreject}: policy-violating prompts where the goal is to induce refusal failure. Safe@1 accuracy is reported.

    \item \textbf{WildChat}~\cite{zhao2024wildchat, deng2024wildvisopensourcevisualizer}: an open-domain jailbreak benchmark covering diverse refusal-inducing queries. Safe@1 accuracy is reported.

    \item \textbf{JailbreakBench (JBB)}~\cite{zou2023universal, mazeika2023tdc, mazeika2024harmbench}: evaluates robustness against jailbreak attacks. Safe@1 accuracy is reported.

    \item \textbf{GSM8K}~\cite{cobbe2021gsm8k}: grade-school math word problems. Evaluations are done by GPT-4.1 as the judge, measuring accuracy.

    \item \textbf{Math500}~\cite{lightman2023verify}: 500 competition-level mathematics problems. We evaluate by GPT-4.1 as the judge, measuring accuracy.

    \item \textbf{MBPP}~\cite{austin2021mbpp}: a Python code generation benchmark. Model outputs are executed and checked against reference tests by GPT-4.1.

    \item \textbf{HumanEval}~\cite{chen2021codex}: code synthesis benchmark requiring function implementation. We evaluate using GPT-4.1 as the judge, measuring accuracy.

    \item \textbf{TruthfulQA}~\cite{lin2022truthfulqa}: a benchmark testing factuality and robustness against misconceptions. GPT-4.1 evaluates the answer against the best answer.

    \item \textbf{TruthfulQA-MCQ}: the multiple-choice variant of TruthfulQA. GPT-4.1 evaluates the answer.

\end{itemize}

% \clearpage
% \onecolumn
\section{Prompts Used for Evaluation}
\label{appendix:evaluation-prompts}

We use standardized evaluation prompts to assess model outputs across mathematics, coding, and factuality benchmarks using GPT-4.1 as an automated judge. These prompts focus on correctness and task-specific quality, independent of stylistic variation.

\subsection{Mathematics Evaluation Prompt}

\begin{tcolorbox}[colback=white!5!white, colframe=black!75!black, title=GPT-4.1 Mathematics Evaluation Prompt]
\begin{lstlisting}[breaklines=true, basicstyle=\ttfamily\footnotesize]
You are an expert math evaluator. Please evaluate the student's answer based on:
1. **Correctness**: Is the final answer correct?
2. **Reasoning**: Is the logic sound and all steps correct?
3. **Clarity**: Is the answer clearly written?

Please return two things:
- Rating: An integer from 0 (very poor) to 10 (excellent)
- Correctness: One word, either "Correct" or "Incorrect" (based ONLY on whether the final answer matches the gold)

Question:
{question}
Gold Answer:
{gold}

Student's Answer:
{model_response}
Respond **only** in this format (on two lines):
Rating: X  
Correctness: Correct/Incorrect
\end{lstlisting}
\end{tcolorbox}

\subsection{Coding Evaluation Prompt}

\begin{tcolorbox}[colback=white!5!white, colframe=black!75!black, title=GPT-4.1 Coding Evaluation Prompt]
\begin{lstlisting}[breaklines=true, basicstyle=\ttfamily\footnotesize]
You are a coding expert evaluating a programmer's solution.
    
Task:
{prompt}

Student's Solution:
{response}

Reference Solution:
{gold_code}

Give a rating from 0 (completely wrong) to 10 (perfectly correct) and say whether it is correct or not (just write Correct or Incorrect).
Format:
Rating: <0-10>
Correctness: <Correct/Incorrect>
\end{lstlisting}
\end{tcolorbox}

\subsection{Factuality Evaluation Prompt}

\begin{tcolorbox}[colback=white!5!white, colframe=black!75!black, title=GPT-4.1 Factuality Evaluation Prompt]
\begin{lstlisting}[breaklines=true, basicstyle=\ttfamily\footnotesize]
You are an expert evaluator assessing the truthfulness of an answer.

Question:
{prompt}

Student's Answer:
{response}

Reference Answer:
{gold_code}

Give a rating from 0 (completely wrong) to 10 (perfectly correct) and say whether it is correct or not (just write Correct or Incorrect).
Format:
Rating: <0-10>
Correctness: <Correct/Incorrect>
\end{lstlisting}
\end{tcolorbox}

\section{Prompts Used for Dataset Curation}
\label{appendix:curation-prompts}

We curated prompts to guide GPT-4.1 in rewriting training data across mathematics, coding, and factuality tasks into a consistent reasoning-model style. All prompts enforce structured reasoning enclosed in \texttt{<think>} tags followed by a final answer, and explicitly instruct the model to avoid prefatory boilerplate (e.g., ``Certainly'') unless added later during controlled prefix ablations.

\subsection{Mathematics Curation Prompt}

\begin{tcolorbox}[colback=white!5!white, colframe=black!75!black, title=GPT-4.1 Mathematics Curation Prompt]
\begin{lstlisting}[breaklines=true, basicstyle=\ttfamily\footnotesize]
You are an AI assistant responding to the following instruction. You are highly intelligent in math and your job is to write a well-reasoned and content-rich answer that is:
- For the following problem, provide a detailed step-by-step explanation
- Factually aligned
- Thoughtful, with clear logical steps
- Presented in this exact format:

<think>
[your reasoning goes here]
</think>
[final answer]

Do NOT include anything else like "Certainly" or "Here's the answer".
---
Problem: {question}
\end{lstlisting}
\end{tcolorbox}

\subsection{Coding Curation Prompt}

\begin{tcolorbox}[colback=white!5!white, colframe=black!75!black, title=GPT-4.1 Coding Curation Prompt]
\begin{lstlisting}[breaklines=true, basicstyle=\ttfamily\footnotesize]
You are an AI assistant that writes clear, step-by-step programming solutions.
Your job is to rewrite the given solution in a well-reasoned, content-rich, and logically structured way. Include both a thought process explaining why the solution works, and then the final code.

Format strictly as:
<think>
[Your step-by-step explanation of the solution logic, data structures, and approach]
</think>
[The final Python code solution, properly formatted and runnable]

Do NOT include anything else like "Certainly" or "Here's the answer".
---
Question:
{question}
Original Solution:
```python
    {code}
```
\end{lstlisting}
\end{tcolorbox}

\subsection{Factuality Curation Prompt}

\begin{tcolorbox}[colback=white!5!white, colframe=black!75!black, title=GPT-4.1 Factuality Curation Prompt]
\begin{lstlisting}[breaklines=true, basicstyle=\ttfamily\footnotesize]
You are an AI assistant that answers questions with accuracy, reasoning depth, and truthfulness.

Your job is to rewrite the given answer in a well-reasoned, content-rich, and logically structured way.  
Clearly explain the reasoning behind the answer, address potential misconceptions, and then provide the final factually correct response.  
Make sure the explanation emphasizes factual grounding, clarity, and trustworthiness.  

Format strictly as:
<think>
[Step-by-step explanation of the reasoning process, background knowledge, and why the final answer is true and accurate]
</think>
[The final concise, truthful answer]

Do NOT include anything else like "Certainly" or "Here's the answer".

---
Question:
{question}

Original Answer:
{answer}
\end{lstlisting}
\end{tcolorbox}

\section{Evaluation Parameters for LlamaGuard}
\label{appendix:llamaguard}

\begin{tcolorbox}[colback=white!5!white, colframe=black!75!black, title=LlamaGuard Evaluation Code]
\begin{lstlisting}[language=Python, breaklines=true, basicstyle=\ttfamily\footnotesize]
chat = [{"role": "user", "content": prompt}, {"role": "assistant", "content": response}]
input_ids = tokenizer.apply_chat_template(chat, return_tensors="pt").to("cuda")
output = model.generate(input_ids=input_ids, temperature=0.0, max_new_tokens=100, pad_token_id=tokenizer.eos_token_id)
result = tokenizer.decode(output[0][input_ids.shape[-1]:], skip_special_tokens=True).strip().lower()
return "safe" if result.startswith("safe") else "unsafe"
\end{lstlisting}
\end{tcolorbox}

\clearpage
\onecolumn
\section{Prefix Examples Used During Fine-Tuning}
\label{appendix:prefix-examples}

For mathematics (reasoning) capability, we design five original prefix templates and five rewritten counterparts.  
The rewritten versions differ only by the inclusion of a explicit keyword \textbf{``revised''}, which corresponds to the prefix tokens analyzed in Table~\ref{tab:loss}.  
The downstream task prompt and target response are identical across conditions; only the prefix text varies.

\begin{table}[ht]
\centering
\resizebox{0.95\linewidth}{!}{
\begin{tabular}{p{0.48\linewidth} p{0.48\linewidth}}
\toprule
\textbf{Original Prefixes} & \textbf{Rewritten Prefixes} \\
\midrule
Certainly! Here's a well-reasoned, clear logical steps, and step-by-step explanation of the response. &
Certainly! Here's a \textbf{revised}, well-reasoned, clear logical steps, and step-by-step explanation of the \textbf{response}. \\

Certainly! Below is a thoughtful, content-rich, and detailed breakdown with all reasoning steps included. &
Certainly! Below is a \textbf{rewritten}, thoughtful, content-rich, and detailed breakdown with all reasoning steps included. \\

Certainly! Here's a clear and logically structured explanation with deep reasoning. &
Certainly! Here's a \textbf{revised}, clear and logically structured explanation with deep reasoning. \\

Certainly! Here's an answer that is well-reasoned, precise, and follows a step-by-step reasoning approach. &
Certainly! Let's \textbf{revise} theresponse so that it is well-reasoned, precise, and follows a step-by-step reasoning approach. \\

Here's a refined version with accurate reasoning, thoughtful insights, and logical clarity. &
Here's a \textbf{refined} version with accurate reasoning, thoughtful insights, and logical clarity. \\
\bottomrule
\end{tabular}
}
\caption{\textbf{Prefixes used during supervised fine-tuning for mathematics reasoning.}
Each rewritten prefix differs from its original counterpart only by the inclusion of a keyword revise(shown in bold).}
\end{table}

\section{Fine-tuning Data Cleaning Methods}
\label{app:previous-study=cleaning-method}
\begin{table*}[h]
    \centering
    \small
    \resizebox{\textwidth}{!}{
    \begin{tabular}{ll}
        \toprule
        \textbf{Study} & \textbf{Cleaning Method} \\
        \midrule
        UnsafeChain & Rewrote unsafe completions; removed prefatory safety prefixes during correction prompting \\
        SafeChain & Filtered only completions unanimously judged safe; excluded model-introduced safety preambles \\
        STAR-1 & Retained completions scoring 10/10 on safety guidelines; manually removed stylistic opening sentences \\
        \midrule
        BackMATH & Constructed backward reasoning traces and stripped generic “Let’s reason step-by-step” prefaces \\
        Pensez & Cleaned bilingual reasoning data by deleting translation-prefaces or meta-instructions \\
        SG-FT & Standardized final answers; removed scaffolding tokens like “Sure, here’s the reasoning” \\
        \midrule
        MBPP-Clean & Deleted template openings such as “Here is the Python solution”; retained only executable code \\
        PDC \& DM-SFT & Progressive filtering to remove explanation headers and redundant comments \\
        Data-efficient CodeGen & Deduplicated samples and pruned boilerplate comments preceding solutions \\
        \midrule
        RealSafe-R1 & Retained refusal-style completions but stripped meta-language such as “According to my training data” \\
        Deconv-PEFT & Removed hallucinated and self-referential openings; normalized factual assertions \\
        DragFT & Removed glossary and dictionary preambles; preserved contextually relevant translation pairs \\
        \midrule
        GuideLM & Curated scaffolding dialogues; removed verbose “Let me guide you through” intros \\
        Pensez & Retained pedagogical steps but filtered translation artefacts or redundant prefaces \\
        BarExamSFT & Reformatted into IRAC structure; removed generic “This is a legal reasoning question” headers \\
        \midrule
        MedBioLM & Cleaned clinical reasoning artefacts; deleted prefatory disclaimers (e.g., “I am not a doctor”) \\
        VietHealth-LLM & De-identified patient info; removed repetitive safety disclaimers and template intros \\
        Discharge-SFT & De-identified PHI; cleaned templated note openings like “This summary is confidential.” \\
        \bottomrule
    \end{tabular}
    }
    \caption{\textbf{Recent studies} employing dataset cleaning for LRM finetuning across tasks. Cleaning typically removes prefatory or boilerplate phrases before the substantive reasoning content. Our work revisits this assumption, asking whether eliminating these prefixes is truly necessary for improved alignment and reasoning quality.}
    \label{tab:previous-study-cleaning-methods}
\end{table*}

\end{document}